%% file: root.tex
\documentclass[letterpaper, 10 pt, conference]{ieeeconf}  

\IEEEoverridecommandlockouts                              

\usepackage{hyperref}       
\usepackage{url}            
\usepackage{booktabs}       
\usepackage{amsfonts}       
\usepackage{nicefrac}       
\usepackage{microtype}      
\usepackage{graphicx}
\usepackage{amsmath} 
\usepackage{amssymb}  
\usepackage{gensymb}
\usepackage{url}
\usepackage{xcolor}
\usepackage{multirow}
\usepackage{hyperref}
\usepackage{paralist}

\newcommand{\p}{\mathbf{p}}
\newcommand{\PP}{\mathbf{P}}
\newcommand{\R}{\mathbf{R}}
\newcommand{\K}{\mathbf{K}}

\newcommand{\U}{\mathbf{U}}
\newcommand{\V}{\mathbf{V}}

\newcommand{\LL}{\mathcal{L}}
\newcommand{\II}{\mathcal{I}}
\newcommand{\DD}{\mathcal{D}}
\newcommand{\MM}{\mathcal{M}}

\usepackage{siunitx}
\usepackage{tabularx, booktabs}
\definecolor{Red}{rgb}{1., 0, 0}
\newcolumntype{Y}{>{\centering\arraybackslash}X}
\newcommand{\ignore}[1]{}

\title{\large \bf
Learning 6-DOF Grasping Interaction via Deep Geometry-aware 3D Representations
}

\author{
  Xinchen Yan$^*$ \quad Jasmine Hsu$^\ddag$ \quad Mohammad Khansari$^\dag$ \quad Yunfei Bai$^\dag$ \quad Arkanath Pathak$^\ddag$ \\
  Abhinav Gupta$^\ddag$ \quad James Davidson$^\ddag$ \quad Honglak Lee$^\ddag$\\
  \thanks{$^*$University of Michigan, during internship with Google Brain.}
  \thanks{$^\ddag$Google, $^\dag$X Inc}  
}

\begin{document}

\maketitle
\thispagestyle{empty}
\pagestyle{empty}

\input{icra_intro}

\input{icra_related_work}
\input{icra_formulation}

\input{icra_experiments}

\input{icra_conclusion}




\input{icra_appendix}





\newpage
\bibliographystyle{abbrv} 
\bibliography{root}

\end{document}

%% file: icra_intro.tex
\begin{abstract}
This paper focuses on the problem of learning 6-DOF grasping with a parallel jaw gripper in simulation. 
Our key idea is constraining and regularizing grasping interaction learning through 3D geometry prediction.
We introduce a deep geometry-aware grasping network (DGGN) that decomposes the learning into two steps.
First, we learn to build mental geometry-aware representation by reconstructing the scene (i.e., 3D occupancy grid) from RGBD input via generative 3D shape modeling.
Second, we learn to predict grasping outcome with its internal geometry-aware representation. 
The learned outcome prediction model is used to sequentially propose grasping solutions via analysis-by-synthesis optimization. 
Our contributions are fourfold: 
(1) To best of our knowledge, we are presenting for the first time a method to learn a 6-DOF grasping net from RGBD input;
(2) We build a grasping dataset from demonstrations in virtual reality with rich sensory and interaction annotations. This dataset includes 101 everyday objects spread across 7 categories, additionally, we propose a data augmentation strategy for effective learning;
(3) We demonstrate that the learned geometry-aware representation leads to about 10$\%$ relative performance improvement over the baseline CNN on grasping objects from our dataset.
(4) We further demonstrate that the model generalizes to novel viewpoints and object instances.
\end{abstract}

\section{Introduction}
Learning to interact with and grasp objects is a fundamental and challenging problem in robot learning that combines perception, motion planning, and control. The problem is challenging because it not only requires understanding geometry (the global shape of an object, the local surface around the interaction space) but it also requires estimating physical properties, such as weight, density, and friction. Furthermore, it requires invariance to illumination, object location, and viewpoint. To handle this, current data-driven approaches~\cite{lenz2015deep,pinto2016supersizing,levine2016learning,mahler2016dex,mahler2017dex} use hundreds of thousands of examples to learn a solution.  

While further scaling may help improve performance of these methods, we postulate shape is core to interaction and that additional shape signals to focus learning will boost performance. The notion of using shape and geometry has been pioneered in grasping research~\cite{goldfeder2009columbia,leon2010opengrasp,bohg2010learning,li2016dexterous,vahrenkamp2016part}.

\begin{figure}[t]
\centering
\includegraphics[width=0.8\linewidth]{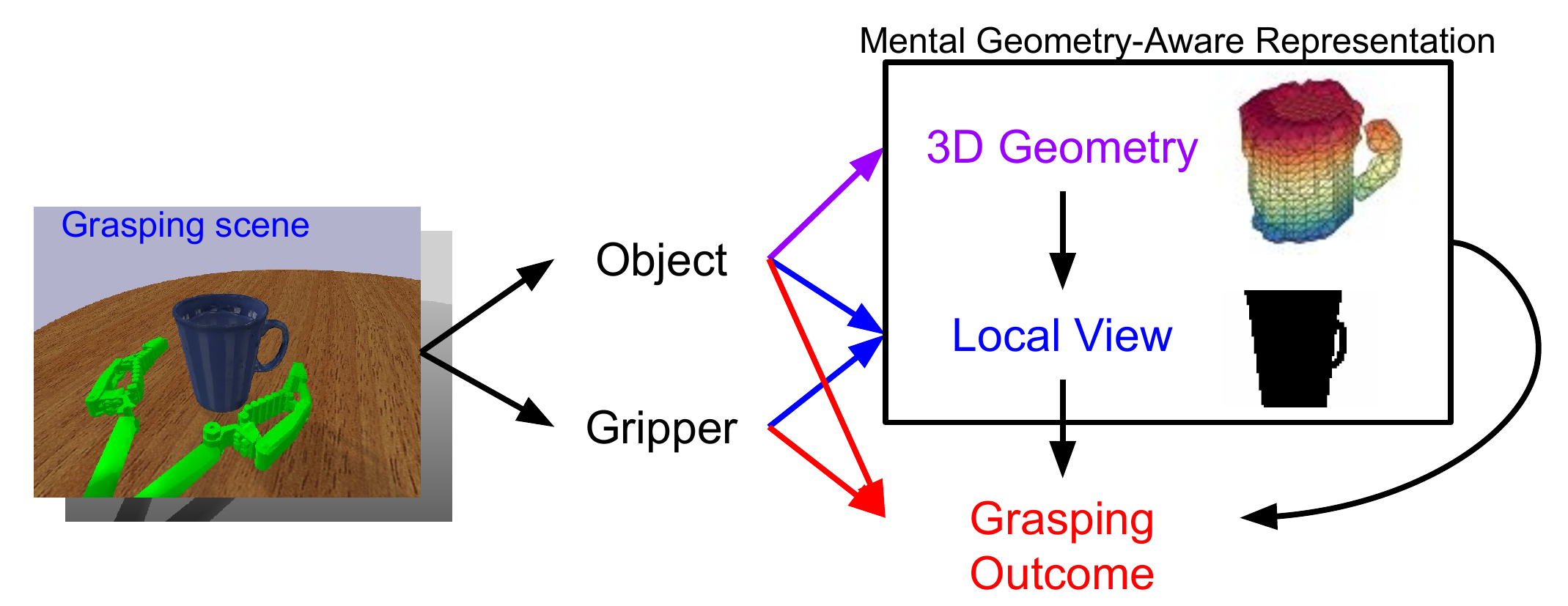} 
\vspace*{-0.1in}
\caption{Learning grasping interactions from demonstrations with deep geometry-aware representations. First, we learn to build mental geometry-aware representation by reconstructing the 3D scene with 2.5D training data. 
Second, we learn to predict grasping outcome with its internal geometry-aware representation.}
\vspace{-5mm}
\label{fig:lfd_intro}
\end{figure}

Inspired by these approaches, we propose the concept of a deep \textbf{geometry-aware} representation (e.g.,~\cite{wu20153d,girdhar2016learning,choy20163d,wu2016learning,maturana2015voxnet,rezende2016unsupervised,yan2016perspective,tulsiani2017multi,godard2016unsupervised,gadelha20163d}) for grasping.
Key to our approach is that we first build a mental representation by \textit{recognizing} and \textit{reconstructing} the 3D geometry of the scene from RGBD input, as demonstrated in Figure~\ref{fig:lfd_intro}. 
With the built-in 3D geometry-aware representation, we can hallucinate a local view of the object's geometric surface from the gripper perspective that will be directly useful for grasping interaction.
In contrast with black-box models that do not have explicit notion of 3D geometry and prior shape-based grasping approaches, our approach has the following features: (1) it performs 3D shape reconstruction as an auxiliary task; (2) it hallucinates the local view using a learning-free physical projection operator; and (3) it explicitly reuses the learned geometry-aware representation for grasping outcome prediction. 

In this work, we design an end-to-end deep geometry-aware grasping network for learning this representation.
Our geometry-aware network has two components: a shape generation network and a grasping outcome prediction network.
The shape generation network learns to recognize and reconstruct the 3D geometry of the scene with an image encoder and voxel decoder.
The image encoder transforms the RGBD input into a high-level geometry representation that involves shape, location, and orientation of the object.
The voxel decoder network takes in the geometry representation and outputs the occupancy grid of the object.
To further hallucinate the local view from gripper perspective, we propose a novel learning-free image projection layer similar to \cite{yan2016perspective,rezende2016unsupervised}.
Building upon the shape generation network,
our grasping outcome prediction network learns to produce a grasping outcome (e.g., success or failure) based on the action (i.e. gripper pose), the current visual state (e.g., object and gripper),
and the learned geometry-aware 3D representation.
%
Unlike our end-to-end multi-objective learning framework, existing data-driven grasping pipelines~\cite{pinto2016supersizing,mahler2016dex,mahler2017dex} can be viewed as models without a shape generation component. 
They require either an additional camera to capture the global object shape or extra processing steps, such as object detection and patch alignment. Furthermore, these methods learn over a constrained grasp space, typically either 3-DOF or 4-DOF.  We relax this constraint to learn fully generalized 6-DOF grasp poses.

We have built a large database consisting of 101 everyday objects with around 150K grasping demonstrations in Virtual Reality with both human and augmented synthetic interactions.
For each object, we collect 10-20 grasping attempts with a parallel jaw gripper from right-handed users.
For each attempt, we record a pre-grasping status which includes the location and orientation of the object and gripper, as well as the grasping outcome (e.g., success or failure given if the object is between the gripper fingers after closing and lifting).
To acquire sufficient data for learning, we generate additional synthetic data by perturbing the gripper location and orientation from human demonstrations using PyBullet~\cite{bulletengine}.
More information about our geometry-aware grasping project can be found at {\color{blue}\underline{\href{https://goo.gl/gPzPhm}{https://goo.gl/gPzPhm}}}.

Our main contributions are summarized below:
\begin{compactitem}
    \item To best of our knowledge, we are presenting for the first time a method to learn a 6-DOF deep grasping neural network from RGBD input.
    \item We build a database with rich visual sensory data and grasping annotations with a virtual reality system and propose a data augmentation strategy for effective learning with only modest amount of human demonstrations. 
    \item We demonstrate that the proposed geometry-aware grasping network is able to learn the shape as well as grasping outcome significantly better than models without notion of geometry.
    \item We demonstrate that the proposed model has advantages in guiding grasping exploration and achieves better generalization to novel viewpoints and novel object instances.
\end{compactitem}

%% file: icra_related_work.tex
\section{Related Work}

A common approach for robotic grasping is to detect the optimal grasping location from 2D or 2.5D visual inputs (RGB or RGBD images, respectively) \cite{saxena2008robotic,montesano2012active,lenz2015deep,pinto2016supersizing,gualtieri2016high,kopicki2016one,osa2016experiments}.
Earlier work~\cite{saxena2008robotic,montesano2012active} studied the planar grasping problem using visual features extracted from 2D sensory input and adopted logistic regression for fitting optimal grasping location with visual features.
Lenz et al.~\cite{lenz2015deep} proposed a two-step detection pipeline (object detection and grasping part detection) with deep neural networks.
Pinto and Gupta~\cite{pinto2016supersizing} built a robotic system for learning grasping from large-scale real-world trial-and-error experiments. In this work, a deep convolutional neural network was trained on 700 hours of robotic grasping data collected from the system.

Fine-grained grasping planning and control often involves 3D modeling of object shape, modeling dynamics of robot hands, and local surface modeling
\cite{goldfeder2009columbia,leon2010opengrasp,johns2016deep,varley2016shape,li2016dexterous,vahrenkamp2016part,mahler2016dex,mahler2017dex}.
Some work focused on analytic modeling of robotic grasps with known object shape information~\cite{goldfeder2009columbia,leon2010opengrasp}.
Varley et al.~\cite{varley2016shape} proposed a shape completion model that reconstructs the 3D occupancy grid for robotic grasping from partial observations, where ground-truth 3D occupancy grid is used during model training. In comparison, our approach does not require full 3D volume supervision for training (e.g., occupancy grid).   
Similar to our work, \cite{bohg2010learning} use a learned shape-context to help predict grasps. Unlike their work, we use the shape to build a virtual global geometric representation along with a local gripper centric model to sequentially propose and evaluate grasp proposals. 
Li et al.~\cite{li2016dexterous} investigated the hand pose estimation in robotic grasping by decoupling contact points and hand configuration with parametrized object shape.
Building upon the compositional aspect of everyday objects, Vahrenkamp et al.~\cite{vahrenkamp2016part} proposed a part-based model for robotic grasping that has better generalization to novel object.
Very recently, effort was also made in building DexNet~\cite{mahler2016dex,mahler2017dex}, a large-scale point cloud database for planar grasping (from top-down).
In addition to general robotic grasping, several recent work investigated the semantic or task-specific grasping~\cite{dang2014semantic,katz2014perceiving,nikandrova2015category}.

In contrast to existing learning frameworks applied to robotic grasping (either top-down grasping or side-grasping), our approach features (1) providing a method to learn a 6D grasping network from RGBD input (2) an end-to-end deep learning framework for generative 3D shape modeling and leveraging it for predictive 6D grasping interaction, and (3) learning-free projection layer that links the 2D observations with 3D object shape which allows for learning the shape representation without explicit 3D volume supervision.

%% file: icra_formulation.tex
\section{Multi-objective framework with geometry-aware representation}

\begin{figure*}[t]
\centering
\hspace*{-0.1in}
\includegraphics[width=1.03\linewidth]{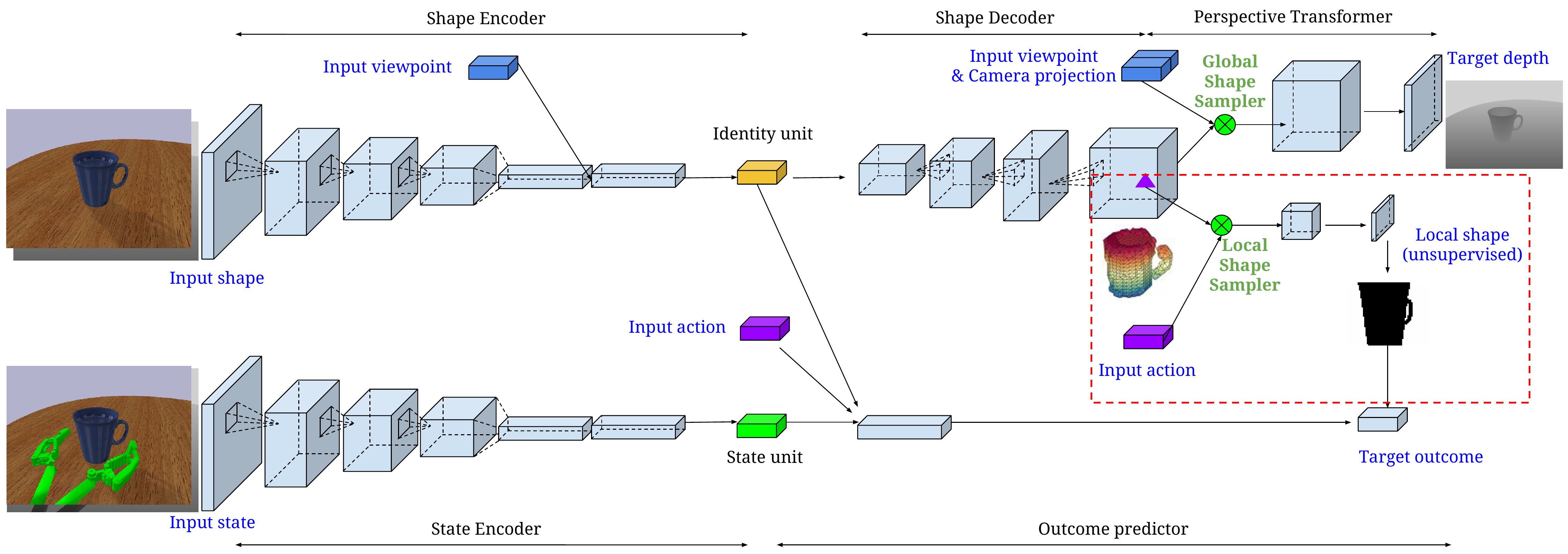} 
\caption{Illustration of DGGN (deep geometry-aware grasping network).
Our DGGN has a shape generation network and an outcome prediction network.
The shape generation network has a 2D CNN encoder, 3D CNN decoder, and a global sampling layer (detailed in Sec.~\ref{sec:depth_projection_layer}).
Our outcome prediction network has a 2D CNN encoder, a local sampling layer (detailed in Sec.~\ref{sec:local_projection_layer}), and a fully-connected prediction network.
}
\vspace{-5mm}
\label{fig:net_arch}
\end{figure*}

In this section, we develop a multi-objective learning framework that performs 3D shape generation and grasping outcome prediction.

\subsection{Learning generative geometry-aware representation from RGBD input}

Being able to \textit{recognize} and \textit{reconstruct} the 3D geometry given RGBD input is a very important step during grasping planning.
In our formulation, we propose a reconstruction of a 3D occupancy grid ~\cite{wu20153d,girdhar2016learning,choy20163d,wu2016learning,rezende2016unsupervised,yan2016perspective,tulsiani2017multi,godard2016unsupervised,gadelha20163d}
that encodes the shape, location, and orientation of the object as our geometry-aware representation.
Previous work generate normalized 3D occupancy grids centered at the origin. 
Our formulated geometry-aware representation differs in that
(1) it takes location and orientation into consideration (the orientation of a novel object is usually undefined);
(2) it is invariant to camera viewpoint and distance (we obtain the same representation from arbitrary camera setting).

Given an RGBD input $\II$ and a corresponding 3D occupancy grid $\V$, the task is to learn a functional mapping $f^V: \II \rightarrow \V$.
Simply following this formulation, previous work~\cite{wu20153d,girdhar2016learning,choy20163d,wu2016learning,maturana2015voxnet} that use 3D supervision obtained reasonable quality in generating normalized 3D volumes by using thousands of shape instances.
However, in our problem setting, these methods would require even more data considering the entangled factors from shape, location, and orientation.

\subsection{Depth supervision with in-network projection layer}
\label{sec:depth_projection_layer}
Recent breakthroughs in reconstructing 3D geometry with 2D supervision~\cite{rezende2016unsupervised,yan2016perspective,tulsiani2017multi,zhou2017unsupervised,godard2016unsupervised,gadelha20163d,fan2016point,tung2017self} suggest that
(1) the quality of reconstructed 3D geometry is as good as previous work with 3D supervision; (2) the learned representation generalizes better to novel settings than previous work with 3D supervision; and (3) learning becomes more efficient with 2D supervision.
Inspired by these findings, we tackle the 3D reconstruction in a weakly supervised manner without explicit 3D shape supervision.
In~\cite{yan2016perspective}, an in-network projection layer is introduced for 3D shape learning from 2D masks (e.g. 2D silhouette of object).
%
Unfortunately, 2D silhouette is usually insufficient supervision signal to reconstruct objects with concave 3D parts (e.g., containers).
For these reasons, we chose to use a depth signal in our shape reconstruction.  Additionally, RGBD sensors are commonly available in most robot platforms.

To enable depth supervision in our shape generation component, we propose a novel in-network OpenGL projection operator that utilizes a 2D depth map $\DD$ as supervision signal for learning to reconstruct the 3D geometry. 
We formulate the projection operation by $f^D: \V \times \PP \rightarrow \DD$ that transforms a 3D shape into a 2D depth map with the camera transformation matrix $\PP$.
Here, the camera transformation matrix decomposes as $\PP = \K[\R; \mathbf{t}]$, where $\K$ is the camera intrinsic matrix, $\R$ is the camera rotation matrix, and $\mathbf{t}$ is the camera translation vector.
In our implementation, we also use a 2D silhouette as an object mask $\MM$ for learning.
Empirically, this additional objective makes the learning stable and efficient.

Following the OpenGL camera transformation standard, 
for each point $\p^s = (x^s, y^s, z^s, 1)$ in 3D world frame,
we compute the corresponding point $\p^n = (x^n, y^n, z^n, 1)$
in the normalized device coordinate system ($-1 \leq x^n, y^n, z^n \leq 1$) using the transformation:
$\p^n \sim \PP \p^s$.
Here, the conversion from depth buffer $z^n$ to real depth $z^e$ is given
by $z^e = f^e(z^n) = -1 / (\alpha * z^n + \beta)$ where $\alpha = \frac{Z_{near}-Z_{far}}{2 Z_{near} Z_{far}}$ and $\beta = \frac{Z_{near} + Z_{far}}{2 Z_{near} Z_{far}}$. Here, $Z_{far}$ and $Z_{near}$ represents the far and near clipping planes of the camera.

Similar to the ``transformer networks" proposed in~\cite{yan2016perspective,jaderberg2015spatial},
our depth projection can be seen as:
(1) performing dense sampling from input volume (in the 3D world frame) to output volume (in normalized device coordinates);
and (2) flattening the 3D spatial output across one dimension.
Again, $j$-th point $(x_j^n, y_j^n, z_j^n)$ in output volume $\U \in \mathbb{R}^{H' \times W' \times D'}$ ($j$-th point is indexed by $[n', m', l']$ in the volume space) and corresponding point $(x_j^s, y_j^s, z_j^s)$ in input volume $\V \in \mathbb{R}^{H\times W\times D}$
are related by the transformation matrix $\PP$.
Here, $(W,H,D)$ and $(W',H',D')$ are the width, height, and depth of the input and output volume, respectively. 
We define the dense sampling step and channel-wise flattening step as follows: 
\begin{align}
		U[n',m',l'] =& \sum_{n=1}^{H} \sum_{m=1}^{W} \sum_{l=1}^{D} V[n,m,l] \max(0,1-|x_j^s-m|) \nonumber \\  
		& \max(0,1-|y_j^s-n|) \max(0,1-|z_j^s-l|) \nonumber \\
		\hat{\MM}[n',m'] =& \max_{l'} U[n',m',l'] \nonumber \\
		\hat{\DD}[n',m'] =& 
			\begin{cases}
				Z_{far},  \text{if } \hat{\MM}[n',m'] = 0\\
				f^e(\frac{2l'}{D'}-1), \\ \text{ where } l'=\arg\min_{l'}(U[n',m',l']>0.5) \\
				Z_{near}, \text{ otherwise}\\
			\end{cases} \nonumber\\
\label{eqn:opengl_transform}
\end{align}
In our implementation, we pre-computed the actual depth $f^e(\frac{2l'}{D'} - 1)$ given the difficulty that $\arg\min$ is not back-propagatable.
As we will see in the following section, the network will be trained to match these predictions $\hat{\MM}$ and $\hat{\DD}$ to the ground-truth ${\MM}$ and ${\DD}$. 
Please note that our in-network projection layer is \textit{learning-free} as it implements the exact ray-tracing algorithm without extra free parameters involved.
We note that the concept of depth projection is also explored in some very recent work~\cite{wu2017marrnet,tewari2017mofa,zhou2017unsupervised}, but their implementations are not exactly the same as our OpenGL projection layer in Eq.~\ref{eqn:opengl_transform}.

\subsection{Viewpoint-invariant geometry-aware representation with multi-view supervision}
Learning to reconstruct 3D geometry from single-view RGBD sensory input is a challenging task in computer vision due to shape ambiguity.
We adopt the shape consistency learning that enforces viewpoint-invariance across multi-view observations~\cite{choy20163d,yan2016perspective,tulsiani2017multi}.
More specifically, we (1) use the averaged identity units from multiple viewpoints as input to shape decoder network and (2) provide multiple projections for supervising the 3D shape reconstruction during training.
Such shape consistency learning encourages an image taken from one viewpoint sharing the same representation with the image taken from another viewpoint.
At testing time, we only provide RGBD input from single viewpoint.
Given a series of $n$ observations $\II_1, \II_2, \cdots, \II_n$ of the scene,
the 3D reconstruction can be formulated as $f^V: \{\II_i\}_{i=1}^n \rightarrow \V$.
Similarly, the projection operator from $i$-th viewpoint is $f^D: \V \times \PP_i \rightarrow \DD_i$, where $\DD_i$ and $\PP_i$ are the depth and camera transformation matrix from corresponding viewpoint, respectively.
Finally, we define the shape reconstruction loss $\LL^{shape}$ in Eq.~\ref{eqn:formulation_multi}.
\begin{equation}
\LL_\theta^{shape}
	= \lambda_\DD \sum_{i=1}^n \LL_\theta^{depth}(\hat{\DD_i}, \DD_i)
	+ \lambda_\MM \sum_{i=1}^n \LL_\theta^{mask}(\hat{\MM_i}, \MM_i)
\label{eqn:formulation_multi}
\end{equation}
Here, $\lambda_\DD$ and $\lambda_\MM$ are the constant coefficients for the depth and mask prediction terms, respectively. 

\subsection{Learning predictive grasping interaction with geometry-aware representation.}
\label{sec:local_projection_layer}

As demonstrated in previous work~\cite{oh2015action,finn2016unsupervised,dosovitskiy2016learning,yang2015weakly,pinto2016curious} that learn interactions from demonstrations,
\textit{prediction} of the future state can be a metric for understanding the physical interaction.
In our grasping setting, we define the RGBD input $\II$ as current state , the 6D pre-grasping parameters $\mathbf{a}$ (position and orientation of the parallel jaw gripper) as  action, and the grasping outcome $l$ (e.g., binary label representing a successful grasp or not) as future state.
The future prediction task can be solved by learning a functional mapping $f_{baseline}^l: \II \times \mathbf{a} \rightarrow l$.
We refer to this method as a baseline grasping interaction prediction model, which has been a basis of several recent state-of-the-art grasping methods using deep learning (e.g., \cite{lenz2015deep,levine2016learning,mahler2017dex}).
These work managed to learn such mapping with either (a) millions of randomly generated grasps, (b) additional view from eye/hand perspective, or (c) additional processing steps such as object detection and image alignment.

In comparison, our geometry-aware model is an end-to-end architecture which constrains its prediction with geometry information.
As we learn to reconstruct the 3D geometry, we argue that the \textit{local surface view} (typically from a wrist camera perspective) can be \textit{directly inferred from our viewpoint-invariant geometry-aware representation} $\hat{\DD}^{local} = f^D(\hat{\V}, \mathbf{P}(\mathbf{a}))$, where
$\hat{V} = f^V(\II)$.
Here, we treat the gripper as a virtual camera with the transformation matrix $\mathbf{P}(\mathbf{a})$ with its world-space coordinates given by the 6D pre-grasping parameters $\mathbf{a}$.
In addition to the \textit{local view}, our geometry-aware representation provides a \textit{global view} of the scene $\V$ that takes a shape prior, location, and orientation of object into consideration. 
Finally, given a current observation $\II$, proposed action $\mathbf{a}$, and inferred 3D shape representation $\V$,
we fit a functional mapping $f_{geometry-aware}^l: \II \times \mathbf{a} \times \V \rightarrow l$, where $l$ is the binary outcome.

\subsection{DGGN: Deep geometry-aware grasping network. }

To implement the two components proposed in the previous sections, we introduce \textbf{DGGN (deep geometry-aware grasping network)} (see Figure~\ref{fig:net_arch}), composed of a shape generation network and an outcome prediction network.
The shape generation network has a 2D convolutional shape encoder and a 3D deconvolutional shape decoder followed by a global projection layer.
Our shape encoder network takes RGBD images of resolution 128 $\times$ 128 and
corresponding 4-by-4 camera view matrices as input; the network outputs identity units as an intermediate representation.
Our shape decoder is a 3D deconvolutional neural network 
that outputs voxels at a resolution of 32 $\times$ 32 $\times$ 32.
We implemented the projection layer (given camera view and projection matrices) that transforms the voxels back 
into foreground object silhouettes and depth maps at an input resolution (128 $\times$ 128).
Here, the purpose of generative pre-training is to learn viewpoint invariant units (e.g., object identity units) 
through object segmentation and depth prediction.
The outcome prediction network has a 2D convolutional state encoder and a fully connected outcome predictor with an additional local shape projection layer. 
Our state encoder takes RGBD input (the pre-grasp scene) of resolution 128 $\times$ 128 and corresponding actions (position and orientation of the gripper end-effector) and outputs state units as intermediate representation.
Our outcome predictor takes both current state (e.g., the pre-grasp scene and gripper action) and geometry features (e.g., viewpoint-invariant global and local geometry from the local projection layer) into consideration.
Note that the local dense-sampling transforms the surface area around the gripper fingers into a foreground silhouette and a depth map at resolution 48 $\times$ 48.
%

%% file: icra_experiments.tex
\section{Experiments}

This section describes our data collection and augmentation process, as well as experimental evaluation on grasping outcome prediction and grasping trials.

\subsection{Dataset collection}

\paragraph{Human demonstrations in VR} We collected grasping demonstrations on seven categories of objects, which include a total of 101 everyday objects.
To collect grasping demonstrations, we set up the HTC Vive system in Virtual Reality (VR)
and assign target objects randomly to five 
right-handed users (three males and two females).
In total, 1597 human grasps are demonstrated, with an average of 15 grasps per object 
(with lowest and highest number of grasps at 7 and 39 for a plate and a wine glass, respectively).
We randomly split 101 objects into three sets (e.g., training, validation and testing) and make sure each set covers the seven categories (70\% for training, 10\% for validation and 20\% for testing). 

\begin{figure}[t]
\centering
\includegraphics[width=1.03\linewidth]{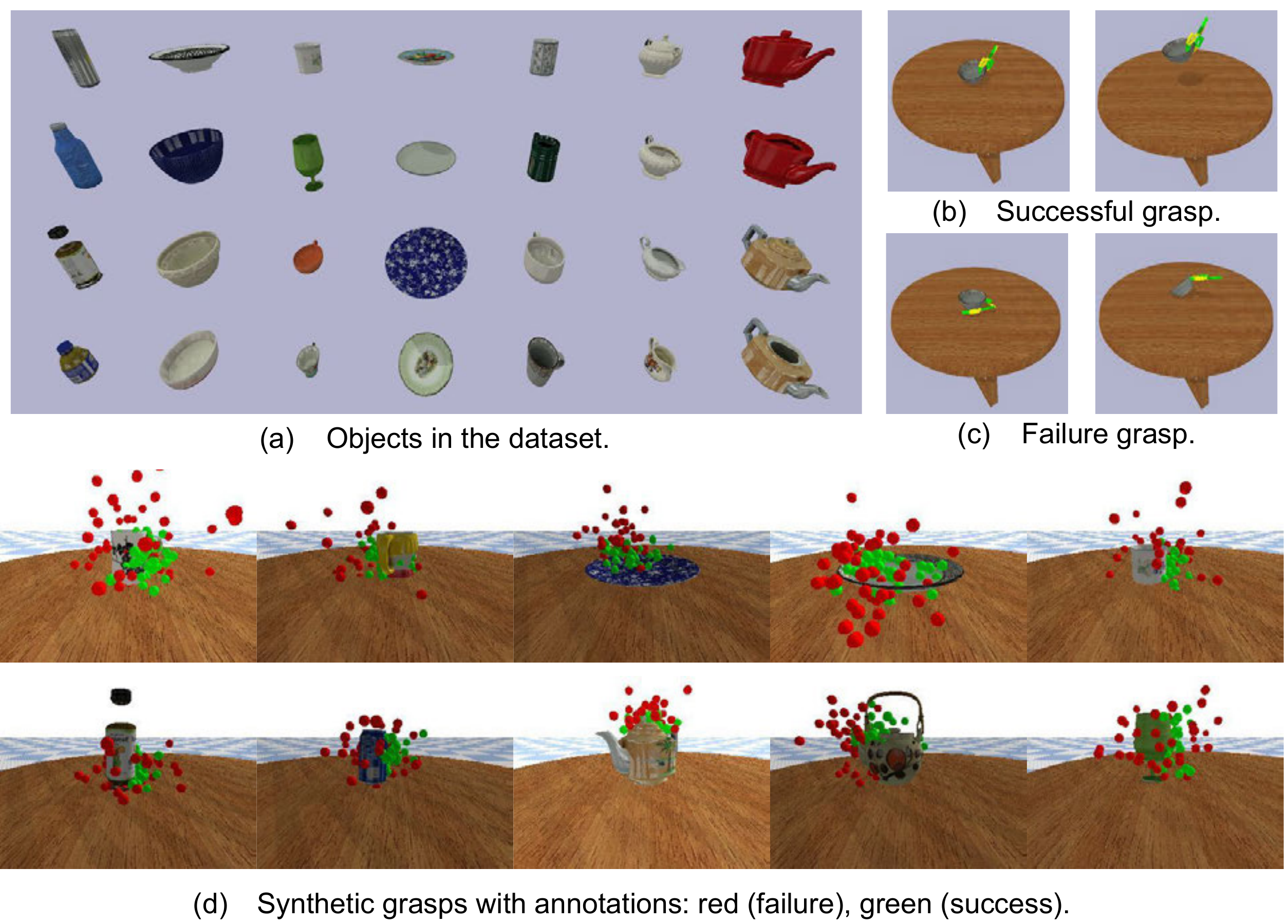}
\caption{Illustrations of our VR-Grasping-101 dataset.}
\label{fig:fig_dataset}
\vspace*{-0.2in}
\end{figure}

\paragraph{Data augmentation}
In order to collect sufficient grasping demonstrations for model training and evaluation, we generate synthetic grasps by perturbing the human demonstrations using PyBullet~\cite{bulletengine}.
This significantly helps in increasing the number of grasps by adding perturbations to the demonstrations.
In total, we collected 150K grasping demonstrations covering 101 objects.
%
%
Figure~\ref{fig:fig_dataset} illustrates examples of objects in the dataset, successful and unsuccessful grasping trials from human demonstrations, and synthetic grasps (visualized by gripper positions) for successful and unsuccessful trials that were generated by this augmentation process.
More details are described in the Appendix.

For each demonstration, we take a snapshot of the pre-grasping scene (e.g., before closing the two gripper fingers).
by randomly setting the camera at a distance (ranging
between 35 centimetres and 45 centimetres).
We draw a camera target position from a normal distribution with its mean as the object center and a desired variance
(in our experiment, we use 3 centimetres as standard deviation).
Furthermore, we set up the camera around the target position from 8 different azimuth angles (with steps of 45 degrees) and adjust the
elevation from 4 different angles (e.g., 15, 30, 45, and 60 degrees).
Finally, we save a state of the scene without a gripper, which is used for shape pre-training; this will be referred to as the static scene throughout the paper.
We include only two elevation angles (e.g., 15 and 45 degrees) in the training set while leaving the rest for evaluation.

\begin{figure*}[t]
\centering
\hspace*{-0.12in}
\includegraphics[width=1.0\linewidth]{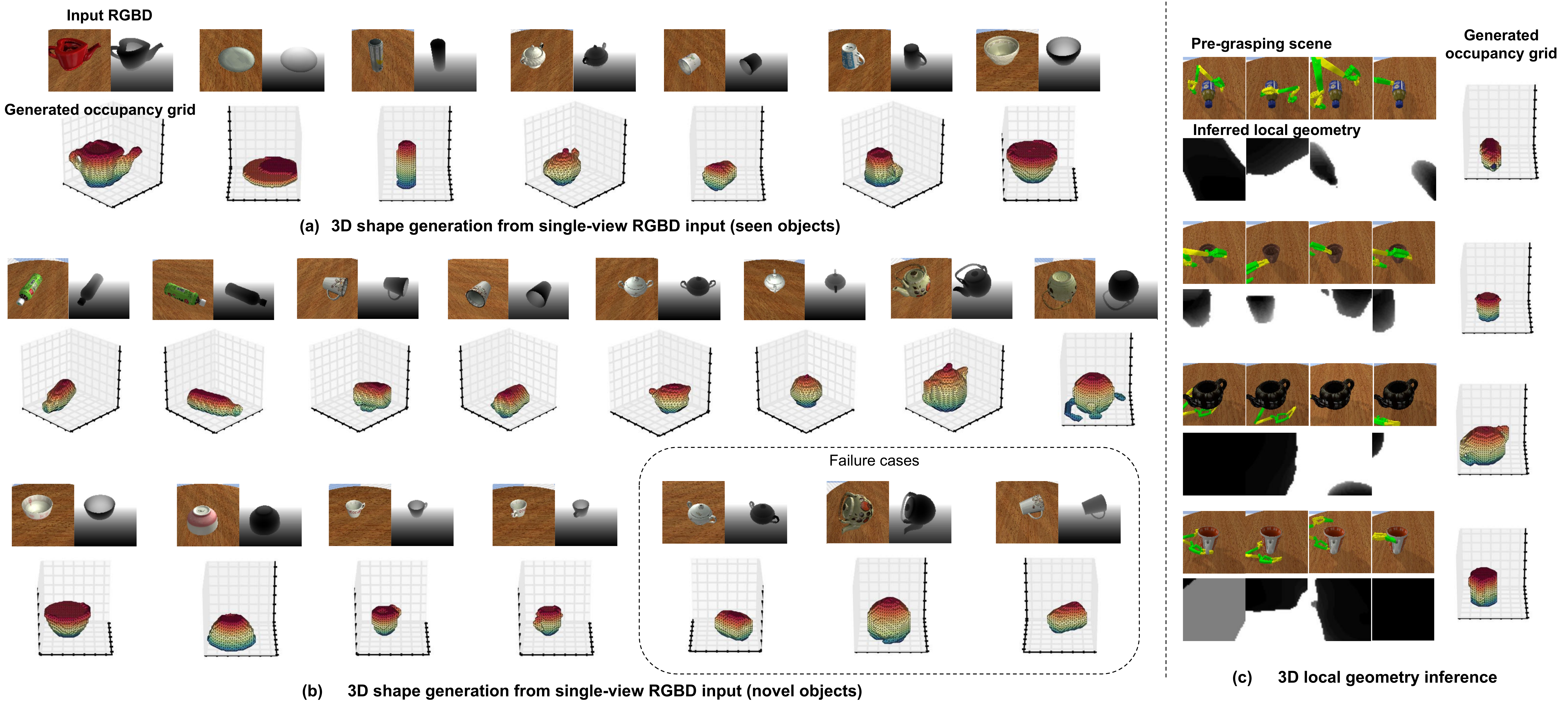}
\caption{Visualization: 3D shape generation from single-view RGBD.
(a) The performance on training (seen) objects. 
(b) The performance on testing (novel) objects.
(c) Local geometry inference from generated occupancy grid.
}
\label{fig:fig_shape_vis}
\end{figure*}

\subsection{Implementation details}

\paragraph{Deep CNN baseline.} We adopt the current data-driven framework as our grasping baseline by removing the shape encoder and shape decoder from our deep geometry-aware grasping model.
This baseline can be interpreted as the grasping quality CNN~\cite{mahler2017dex} without an additional view from a top-down camera.
We trained the model using the ADAM optimizer with a learning rate of $10^{-5}$ for 200K iterations and a mini-batch of size of 4. As an ablation study, we added view and static scene as an additional input channel on top of the baseline model but didn't observe significant improvements.
%

\paragraph{Training DGGN.} We adopted a two-stage training procedure: 
First, we pre-trained the shape generation model (shape encoder and shape decoder) using the ADAM optimizer with a learning rate of $10^{-5}$ for 400K iterations and a mini-batch of size of 4.
In each batch, we sample 4 random viewpoints for the purpose of multi-view supervision in the training time.
We observed that this setting led to a more stable shape generation performance compared to single-view training.
In addition, we used ${L}_1$ loss for foreground depth prediction and ${L}_2$ loss for silhouette prediction with coefficients $\lambda_\mathcal{D}=0.5$ and $\lambda_\mathcal{M}=10.0$.
In the second stage, we fine-tuned the state encoder and outcome predictor using the ADAM optimizer with a learning rate of $3*10^{-6}$ for 200K iterations and a mini-batch of size of 4.
We used cross-entropy as our objective function since the grasping prediction is formulated as a binary classification task.

In our experiments, all the models are trained using 20 GPU workers and 32 parameter servers with asynchronized updates. 
Both baseline and our geometry-aware model adopt convolutional encoder-decoder architecture with residual connections. 
The bottleneck layer (e.g., the identity unit in the geometry-aware model) is a $768$ dimensional vector.

\subsection{Visualization: 3D shape generation}

We evaluate the quality of the shape generation model by visualizing the geometry representations through the shape encoder and decoder network.
In our evaluations, we used single-view RGBD input and corresponding camera view matrix as input to the network.
As shown in Figure~\ref{fig:fig_shape_vis}(a), our shape generation model is able to generate a detailed 3D occupancy grid from single-view input without 3D supervision during training.
As shown in Figure~\ref{fig:fig_shape_vis}(b), our model demonstrates reasonable generalization quality even on novel object instances.

\begin{table*}[t!]
\centering
\begin{tabular}{l||c|c|c|c|c|c|c||c}
\hline
Method $/$ Category & bottle & bowl & cup & plate & mug & sugarbowl & teapot & all \\
\hline\hline
baseline CNN (15) & 72.81 & 73.36 & 73.26 & 66.92 & 72.23 & 70.45 & 66.13 & 71.42\\
\hline
our DGGN (15) & \textbf{78.83} & \textbf{79.32} & \textbf{77.60} & \textbf{68.88} & \textbf{78.25} & \textbf{76.09} & \textbf{73.69} & \textbf{76.55} \\
\hline\hline
baseline CNN (45) & 71.02 & 74.16 & 73.50 & 63.31 & 74.23 & 72.70 & 64.19 & 71.32\\
\hline
our DGGN (45) & \textbf{78.77} & \textbf{80.63} & \textbf{78.06} & \textbf{70.13} & \textbf{79.29} & \textbf{77.52} & \textbf{72.88} & \textbf{77.25}\\
\hline	
\end{tabular}
\caption{Grasping Outcome prediction accuracy from seen elevation angles.}
\label{tab:outcome_eval_seen_angle}
\begin{tabular}{l||c|c|c|c|c|c|c||c}
\hline
Method $/$ Category & bottle & bowl & cup & plate & mug & sugarbowl & teapot & all \\
\hline\hline
baseline CNN (30) & 71.15 & 72.98 & 71.65 & 61.90 & 71.01 & 70.06 & 61.88 & 69.50\\
\hline	
DGGN (30) & \textbf{79.17} & \textbf{77.71} & \textbf{77.23} & \textbf{67.00} & \textbf{75.95} & \textbf{75.06} & \textbf{70.66} & \textbf{75.27}\\
\hline\hline
baseline CNN (60) & 68.45 & 73.05 & 72.50 & 61.27 & 74.40 & 71.30 & 63.25 & 70.18\\	
\hline
DGGN (60) & \textbf{77.40} & \textbf{78.52} & \textbf{76.24} & \textbf{68.13} & \textbf{79.39} & \textbf{76.15} & \textbf{70.34} & \textbf{75.76}\\
\hline	
\end{tabular}
\caption{Grasping Outcome prediction accuracy from novel elevation angles.}
\label{tab:outcome_eval_novel_angle}
\vspace*{-0.3in}
\end{table*}

\paragraph{Analysis: local geometry inference via projection.}
One advantage of our shape generation component is that we can obtain additional local geometry information (see the red-dashed box in Figure~\ref{fig:net_arch}(c))
from our geometry-aware representation.
This is the key difference between our work and the related work that require additional camera from the gripper.
With 3D geometry as part of the intermediate representation, we hallucinate the local geometry by running a projection from the gripper's perspective (i.e., simply treat the gripper as another virtual camera).
To further understand the advantages of our shape generation component, we visualized the intermediate local geometry projected from generated 3D occupancy grid.
As shown in Figure~\ref{fig:fig_shape_vis}(c), our shape generation component provides accurate local geometry estimation that is useful for grasping outcome prediction.

\subsection{Model evaluation: Grasping outcome prediction}

To evaluate the actual advantages in grasping outcome prediction from our modeling, 
we computed the average classification accuracy over 30K demonstrations from novel object instances (from testing set) with diverse observation viewpoints.
For each human demonstration, we generated 100 synthetic grasps through perturbation (among which 50\% of them are success grasps) and computed the average accuracy on 100 grasps (i.e., random guess achieves 50\% accuracy).
To investigate the model performance due to viewpoint changes, we repeat the evaluation experiment for four different elevation angles (e.g, 15, 30, 45, and 60 degrees).
We use parallel computing resources ($500$ machines) during evaluation and the entire evaluation took about $1$ day.
The results are summarized in Table~\ref{tab:outcome_eval_seen_angle} 
and Table~\ref{tab:outcome_eval_novel_angle}.
Overall, the deep geometry-aware model consistently outperforms the deep CNN baseline in grasping outcome classification.
As we can see, ``teapot'' and ``plate'' are comparatively more challenging
categories for outcome prediction, since ``teapot'' has irregular shape parts (e.g., tip and handle) and ``plate'' has a fairly flat shape.
When it comes to novel elevation angles (e.g., compare Table~\ref{tab:outcome_eval_seen_angle} and Table~\ref{tab:outcome_eval_novel_angle}), our deep geometry-aware model is less affected, especially in categories such as ``teapot'' and ``plate'' where viewpoint-invariant shape understanding is crucial.

\begin{table*}[t]
\centering
\begin{tabular}{l||c|c|c|c|c|c|c||c}
\hline
Method $/$ Category & bottle & bowl & cup & plate & mug & sugarbowl & teapot & all \\
\hline\hline
baseline CNN + CEM & 48.60 & 64.28 & 55.44 & 45.99 & 61.00 & 53.97 & 63.08 & 55.85\\
\hline
our DGGN + CEM & \textbf{56.73} & \textbf{68.84} & \textbf{60.31} & \textbf{50.09} & \textbf{67.21} & \textbf{59.87} & \textbf{69.22} & \textbf{61.46}\\
\hline\hline
rel. improvement (\%) & 16.72 & 7.09 & 8.77 & 8.92 & 10.18 & 10.92 & 9.73 & 10.03 \\
\hline
\end{tabular}
\caption{Grasping planning on novel objects: success rate by optimizing for up to 20 steps.}
\label{tab:grasping_opt_novel}
\vspace*{-0.3in}
\end{table*}

\begin{figure}[ht]
\centering
\includegraphics[width=1.0\linewidth]{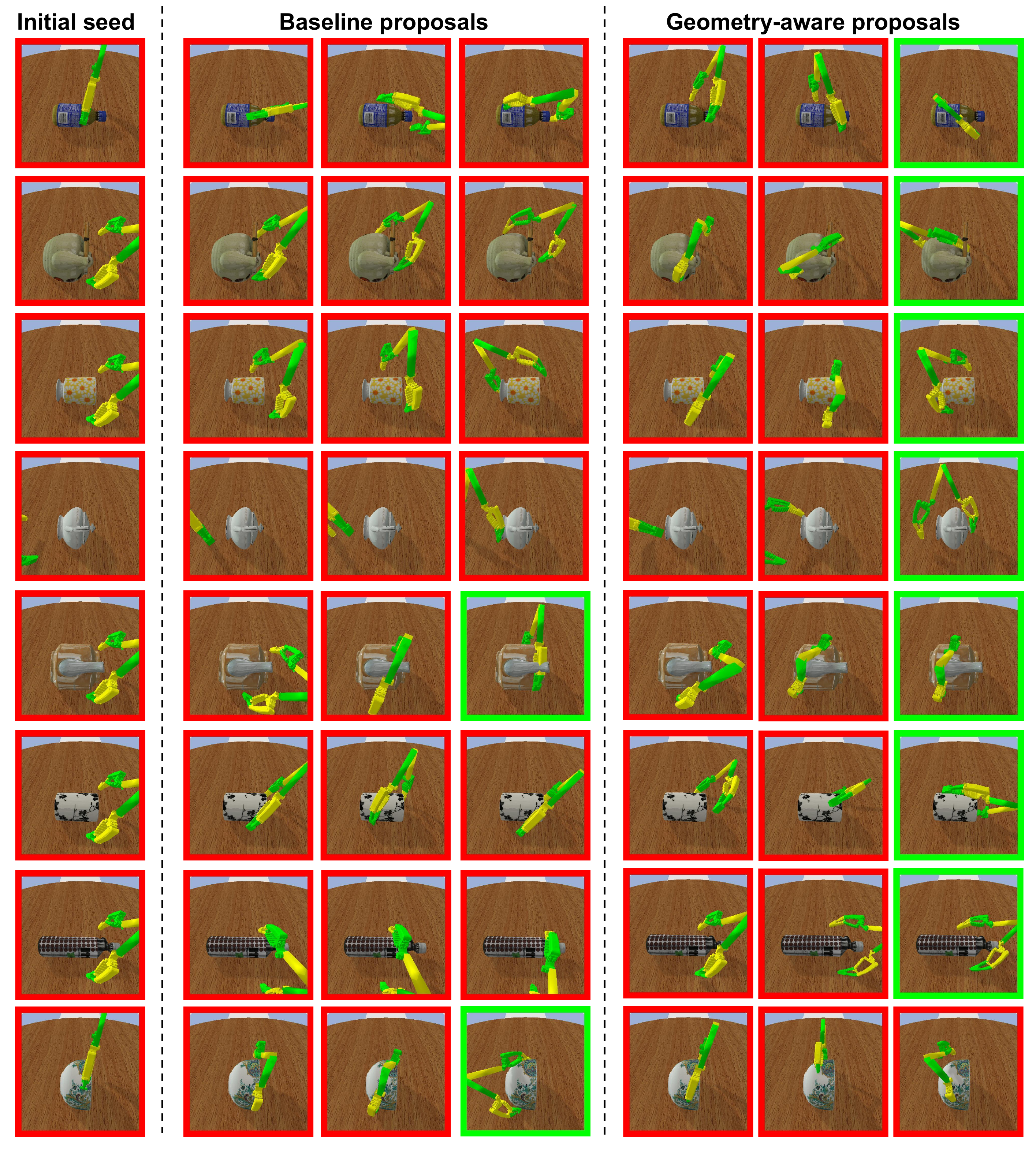}
\caption{Visualization: grasping optimization with CEM based on the grasping prediction output. In each row, we selected three representative steps in grasping optimization (in sequential order from left to right).
Red box represents a failure grasp while green box represents a successful grasp.}
\label{fig:fig_grasping_vis}
\end{figure}

\subsection{Application: Analysis-by-synthesis grasping planning.}

As we improve the classification accuracy over the grasping outcome, a natural question is whether this improvement can be used to guide better grasping planning.
Given a grasping proposal (defined as target gripper pose) seed, we conducted grasping planning by sequentially adjusting the grasping pose guided by our deep grasping network until a grasp success.
In each optimization step, we performed cross-entropy method (CEM)~\cite{rubinstein2004cem,levine2016learning} as follows.
(1) We initialized with a failure grasp in order to force the model to find better grasping pose.
(2) To obtain the gradient direction in the 6D space, we sample 10 random directions and selected the top one based on the score returned by the neural network (output of outcome predictor). We repeat the iterations until success (we set an upper bound of 20 steps).
We conducted the same grasping explore evaluation for both the baseline CNN and our deep geometry-aware model.
To account for the variations in observation viewpoints and initial seeds, we repeat the evaluation for eight times per testing demonstration in our dataset and reported the average success rate after 20 iterations (marked as failure only if there is no success in 20 steps).
As shown in Table~\ref{tab:grasping_opt_novel}, CEM guided our geometry-aware model performance consistently better than the baseline CNN model.
We believe the improved performance comes from the explicit modeling of the 3D geometry as intermediate representation in our deep geometry-aware model.
Our model achieved the most significant improvement in the ``bottle'' category, since a bottle shape is relatively easy to reconstruct.
%
Our improvement in the ``bowl'' category is less significant, partly due to the difficulty of predicting its concave shape in novel object instances. 
Figure~\ref{fig:fig_grasping_vis} demonstrates example grasping planning trajectories on different objects. 
The baseline CNN is less robust compared to our deep geometry-aware model,
which is more likely to transit from one side of the object to the other side with a clear notion of 3D geometry.

%% file: icra_conclusion.tex
\section{Conclusions and Future Work}
In this work, we studied the problem of learning the grasping interaction with deep geometry-aware representation. 
We proposed a deep geometry-aware network that performs shape generation as well as grasping outcome prediction with a learning-free physical projection layer.
Compared to the CNN baseline, experimental results demonstrated improved performance in outcome prediction thanks to generative shape modeling.
Guided by the geometry-aware representation, we obtained better planning via analysis-by-synthesis grasping optimization.

We believe the proposed deep geometry-aware grasping framework has many potentials in advancing robot learning in general.
One interesting future direction is to apply the learned geometry-aware representation to perform tasks using other types of hands (e.g., hands with very different kinematics).
In addition, we would like to explore some alternative model designs (e.g., learn to grasp without the auxiliary state encoder) such that the learned geometry-aware representation might be easily adapted to other domains (e.g., real robot setup).

%% file: icra_appendix.tex

\section*{Appendix: Details of Synthetic data generation}
We take advantage of the Pybullet simulator~\cite{bulletengine} by switching between two modes: simulation and log playback.
%
%
In essence, we use the following protocols to generate additional grasps:
\begin{itemize}
\item Start the demonstration log playback in the Bullet physics engine.
\item Pause the log playback once a grasp is detected (pre-grasp state).
\item Store the current scene state (position and orientation of all objects).
\item Repeat random grasping exploration 100 times.
(1) Draw a new grasp pose from a normal distribution with its mean being the
value of the demonstrated grasp pose and a desired variance (in our experiment we
use 5 centimetres as standard deviation for position and 20 Euler degrees as
standard deviation for orientation).
(2) Switch to simulation mode, open the gripper finger, place it at the new drawn
random pose, close the gripper, and lift it.
(3) Check whether the object is still between gripper fingers.
Based on the outcome, we add a new pose to the list of successful
or failed grasps (see Figure~\ref{fig:fig_dataset}(b)(c)).
(4) Reset the simulation environment to the previously stored state.
\item Resume log playback until next grasp is detected.
\end{itemize}
With this protocol, we collected a total of 150K grasping synthetic grasps based on human demonstrations.
As shown in Figure~\ref{fig:fig_dataset}(d), we visualize the gripper positions using colored dots (we omit the gripper orientations): green ones representing successful grasps and red ones representing failure grasps.


%% file: root.bbl
\begin{thebibliography}{10}

\bibitem{bohg2010learning}
J.~Bohg and D.~Kragic.
\newblock Learning grasping points with shape context.
\newblock {\em Robotics and Autonomous Systems}, 58(4):362--377, 2010.

\bibitem{choy20163d}
C.~B. Choy, D.~Xu, J.~Gwak, K.~Chen, and S.~Savarese.
\newblock 3d-r2n2: A unified approach for single and multi-view 3d object
  reconstruction.
\newblock In {\em European Conference on Computer Vision}, pages 628--644.
  Springer, 2016.

\bibitem{bulletengine}
E.~Coumans, Y.~Bai, and J.~Hsu.
\newblock Pybullet physics engine.
\newblock \url{http://pybullet.org}.

\bibitem{dang2014semantic}
H.~Dang and P.~K. Allen.
\newblock Semantic grasping: planning task-specific stable robotic grasps.
\newblock {\em Autonomous Robots}, 37(3):301--316, 2014.

\bibitem{dosovitskiy2016learning}
A.~Dosovitskiy and V.~Koltun.
\newblock Learning to act by predicting the future.
\newblock {\em arxiv preprint: 1611.01779}, 2016.

\bibitem{fan2016point}
H.~Fan, H.~Su, and L.~Guibas.
\newblock A point set generation network for 3d object reconstruction from a
  single image.
\newblock In {\em CVPR}, 2017.

\bibitem{finn2016unsupervised}
C.~Finn, I.~Goodfellow, and S.~Levine.
\newblock Unsupervised learning for physical interaction through video
  prediction.
\newblock In {\em Advances in Neural Information Processing Systems}, pages
  64--72, 2016.

\bibitem{gadelha20163d}
M.~Gadelha, S.~Maji, and R.~Wang.
\newblock 3d shape induction from 2d views of multiple objects.
\newblock {\em arXiv preprint arXiv:1612.05872}, 2016.

\bibitem{girdhar2016learning}
R.~Girdhar, D.~F. Fouhey, M.~Rodriguez, and A.~Gupta.
\newblock Learning a predictable and generative vector representation for
  objects.
\newblock In {\em European Conference on Computer Vision}, pages 484--499.
  Springer, 2016.

\bibitem{godard2016unsupervised}
C.~Godard, O.~Mac~Aodha, and G.~J. Brostow.
\newblock Unsupervised monocular depth estimation with left-right consistency.
\newblock In {\em CVPR}, 2016.

\bibitem{goldfeder2009columbia}
C.~Goldfeder, M.~Ciocarlie, H.~Dang, and P.~K. Allen.
\newblock The columbia grasp database.
\newblock In {\em Robotics and Automation, 2009. ICRA'09. IEEE International
  Conference on}, pages 1710--1716. IEEE, 2009.

\bibitem{gualtieri2016high}
M.~Gualtieri, A.~ten Pas, K.~Saenko, and R.~Platt.
\newblock High precision grasp pose detection in dense clutter.
\newblock In {\em Intelligent Robots and Systems (IROS), 2016 IEEE/RSJ
  International Conference on}, pages 598--605. IEEE, 2016.

\bibitem{jaderberg2015spatial}
M.~Jaderberg, K.~Simonyan, A.~Zisserman, et~al.
\newblock Spatial transformer networks.
\newblock In {\em Advances in Neural Information Processing Systems}, pages
  2017--2025, 2015.

\bibitem{johns2016deep}
E.~Johns, S.~Leutenegger, and A.~J. Davison.
\newblock Deep learning a grasp function for grasping under gripper pose
  uncertainty.
\newblock In {\em Intelligent Robots and Systems (IROS), 2016 IEEE/RSJ
  International Conference on}, pages 4461--4468. IEEE, 2016.

\bibitem{katz2014perceiving}
D.~Katz, A.~Venkatraman, M.~Kazemi, J.~A. Bagnell, and A.~Stentz.
\newblock Perceiving, learning, and exploiting object affordances for
  autonomous pile manipulation.
\newblock {\em Autonomous Robots}, 37(4):369--382, 2014.

\bibitem{kopicki2016one}
M.~Kopicki, R.~Detry, M.~Adjigble, R.~Stolkin, A.~Leonardis, and J.~L. Wyatt.
\newblock One-shot learning and generation of dexterous grasps for novel
  objects.
\newblock {\em The International Journal of Robotics Research}, 35(8):959--976,
  2016.

\bibitem{lenz2015deep}
I.~Lenz, H.~Lee, and A.~Saxena.
\newblock Deep learning for detecting robotic grasps.
\newblock {\em The International Journal of Robotics Research},
  34(4-5):705--724, 2015.

\bibitem{leon2010opengrasp}
B.~Le{\'o}n, S.~Ulbrich, R.~Diankov, G.~Puche, M.~Przybylski, A.~Morales,
  T.~Asfour, S.~Moisio, J.~Bohg, J.~Kuffner, et~al.
\newblock Opengrasp: A toolkit for robot grasping simulation.

\bibitem{levine2016learning}
S.~Levine, P.~Pastor, A.~Krizhevsky, J.~Ibarz, and D.~Quillen.
\newblock Learning hand-eye coordination for robotic grasping with deep
  learning and large-scale data collection.
\newblock {\em The International Journal of Robotics Research}, page
  0278364917710318.

\bibitem{li2016dexterous}
M.~Li, K.~Hang, D.~Kragic, and A.~Billard.
\newblock Dexterous grasping under shape uncertainty.
\newblock {\em Robotics and Autonomous Systems}, 75:352--364, 2016.

\bibitem{mahler2017dex}
J.~Mahler, J.~Liang, S.~Niyaz, M.~Laskey, R.~Doan, X.~Liu, J.~A. Ojea, and
  K.~Goldberg.
\newblock Dex-net 2.0: Deep learning to plan robust grasps with synthetic point
  clouds and analytic grasp metrics.
\newblock {\em arxiv preprint: 1703.09312}, 2017.

\bibitem{mahler2016dex}
J.~Mahler, F.~T. Pokorny, B.~Hou, M.~Roderick, M.~Laskey, M.~Aubry,
  K.~Kohlhoff, T.~Kr{\"o}ger, J.~Kuffner, and K.~Goldberg.
\newblock Dex-net 1.0: A cloud-based network of 3d objects for robust grasp
  planning using a multi-armed bandit model with correlated rewards.
\newblock In {\em IEEE International Conference on Robotics and Automation
  (ICRA)}, pages 1957--1964. IEEE, 2016.

\bibitem{maturana2015voxnet}
D.~Maturana and S.~Scherer.
\newblock Voxnet: A 3d convolutional neural network for real-time object
  recognition.
\newblock In {\em Intelligent Robots and Systems (IROS), 2015 IEEE/RSJ
  International Conference on}, pages 922--928. IEEE, 2015.

\bibitem{montesano2012active}
L.~Montesano and M.~Lopes.
\newblock Active learning of visual descriptors for grasping using
  non-parametric smoothed beta distributions.
\newblock {\em Robotics and Autonomous Systems}, 60(3):452--462, 2012.

\bibitem{nikandrova2015category}
E.~Nikandrova and V.~Kyrki.
\newblock Category-based task specific grasping.
\newblock {\em Robotics and Autonomous Systems}, 70:25--35, 2015.

\bibitem{oh2015action}
J.~Oh, X.~Guo, H.~Lee, R.~L. Lewis, and S.~Singh.
\newblock Action-conditional video prediction using deep networks in atari
  games.
\newblock In {\em Advances in Neural Information Processing Systems}, pages
  2863--2871, 2015.

\bibitem{osa2016experiments}
T.~Osa, J.~Peters, and G.~Neumann.
\newblock Experiments with hierarchical reinforcement learning of multiple
  grasping policies.
\newblock In {\em International Symposium on Experimental Robotics}, pages
  160--172. Springer, 2016.

\bibitem{pinto2016curious}
L.~Pinto, D.~Gandhi, Y.~Han, Y.-L. Park, and A.~Gupta.
\newblock The curious robot: Learning visual representations via physical
  interactions.
\newblock In {\em European Conference on Computer Vision}, pages 3--18.
  Springer, 2016.

\bibitem{pinto2016supersizing}
L.~Pinto and A.~Gupta.
\newblock Supersizing self-supervision: Learning to grasp from 50k tries and
  700 robot hours.
\newblock In {\em Robotics and Automation (ICRA), 2016 IEEE International
  Conference on}, pages 3406--3413. IEEE, 2016.

\bibitem{rezende2016unsupervised}
D.~J. Rezende, S.~A. Eslami, S.~Mohamed, P.~Battaglia, M.~Jaderberg, and
  N.~Heess.
\newblock Unsupervised learning of 3d structure from images.
\newblock In {\em Advances In Neural Information Processing Systems}, pages
  4997--5005, 2016.

\bibitem{rubinstein2004cem}
R.~Rubinstein and D.~Kroese.
\newblock The cross-entropy method: A unified approach to combinatorial
  optimization, monte-carlo simulation, and machine learning.
\newblock 2004.

\bibitem{saxena2008robotic}
A.~Saxena, J.~Driemeyer, and A.~Y. Ng.
\newblock Robotic grasping of novel objects using vision.
\newblock {\em The International Journal of Robotics Research}, 27(2):157--173,
  2008.

\bibitem{tewari2017mofa}
A.~Tewari, M.~Zollh{\"o}fer, H.~Kim, P.~Garrido, F.~Bernard, P.~Perez, and
  C.~Theobalt.
\newblock Mofa: Model-based deep convolutional face autoencoder for
  unsupervised monocular reconstruction.
\newblock In {\em The IEEE International Conference on Computer Vision (ICCV)},
  volume~2, 2017.

\bibitem{tulsiani2017multi}
S.~Tulsiani, T.~Zhou, A.~A. Efros, and J.~Malik.
\newblock Multi-view supervision for single-view reconstruction via
  differentiable ray consistency.
\newblock In {\em CVPR}, 2017.

\bibitem{tung2017self}
H.-Y. Tung, H.-W. Tung, E.~Yumer, and K.~Fragkiadaki.
\newblock Self-supervised learning of motion capture.
\newblock In {\em Advances in Neural Information Processing Systems}, pages
  5242--5252, 2017.

\bibitem{vahrenkamp2016part}
N.~Vahrenkamp, L.~Westkamp, N.~Yamanobe, E.~E. Aksoy, and T.~Asfour.
\newblock Part-based grasp planning for familiar objects.
\newblock In {\em Humanoid Robots (Humanoids), 2016 IEEE-RAS 16th International
  Conference on}, pages 919--925. IEEE, 2016.

\bibitem{varley2016shape}
J.~Varley, C.~DeChant, A.~Richardson, A.~Nair, J.~Ruales, and P.~Allen.
\newblock Shape completion enabled robotic grasping.
\newblock {\em arxiv preprint: 1609.08546}, 2016.

\bibitem{wu2017marrnet}
J.~Wu, Y.~Wang, T.~Xue, X.~Sun, B.~Freeman, and J.~Tenenbaum.
\newblock Marrnet: 3d shape reconstruction via 2.5 d sketches.
\newblock In {\em Advances In Neural Information Processing Systems}, pages
  540--550, 2017.

\bibitem{wu2016learning}
J.~Wu, C.~Zhang, T.~Xue, B.~Freeman, and J.~Tenenbaum.
\newblock Learning a probabilistic latent space of object shapes via 3d
  generative-adversarial modeling.
\newblock In {\em Advances in Neural Information Processing Systems}, pages
  82--90, 2016.

\bibitem{wu20153d}
Z.~Wu, S.~Song, A.~Khosla, F.~Yu, L.~Zhang, X.~Tang, and J.~Xiao.
\newblock 3d shapenets: A deep representation for volumetric shapes.
\newblock In {\em Proceedings of the IEEE Conference on Computer Vision and
  Pattern Recognition}, pages 1912--1920, 2015.

\bibitem{yan2016perspective}
X.~Yan, J.~Yang, E.~Yumer, Y.~Guo, and H.~Lee.
\newblock Perspective transformer nets: Learning single-view 3d object
  reconstruction without 3d supervision.
\newblock In {\em Advances in Neural Information Processing Systems}, pages
  1696--1704, 2016.

\bibitem{yang2015weakly}
J.~Yang, S.~E. Reed, M.-H. Yang, and H.~Lee.
\newblock Weakly-supervised disentangling with recurrent transformations for 3d
  view synthesis.
\newblock In {\em Advances in Neural Information Processing Systems}, pages
  1099--1107, 2015.

\bibitem{zhou2017unsupervised}
T.~Zhou, M.~Brown, N.~Snavely, and D.~G. Lowe.
\newblock Unsupervised learning of depth and ego-motion from video.
\newblock In {\em CVPR}, 2017.

\end{thebibliography}
